\documentclass[smallextended]{svjour3}

\smartqed  
\usepackage{booktabs}
\usepackage{enumitem}
\usepackage{longtable}
\usepackage{graphicx}
\usepackage{array}
\usepackage{dcolumn}
\usepackage{amsfonts}
\usepackage{amssymb}
\usepackage{xspace}
\usepackage{xfrac}
\usepackage{array}
\usepackage{multirow}
\usepackage{xcolor}
\usepackage{footnote}
\usepackage{url}
\usepackage{makeidx}
\usepackage{epsfig}
\usepackage{amsmath}
\usepackage{multicol}
\usepackage{mathtools}
\usepackage[misc]{ifsym}
\usepackage[ruled,linesnumbered]{algorithm2e}

\usepackage{caption}
\usepackage{subcaption}
\begin{document}

\title{Early Anomaly Detection in Time Series}
\subtitle{A Hierarchical Approach for Predicting Critical Health Episodes}

\titlerunning{Early Anomaly Detection in Time Series}

\author{Vitor Cerqueira, Luis Torgo, and Carlos~Soares}

\authorrunning{V. Cerqueira et al.}

\institute{Vitor Cerqueira (\Letter) \at
         LIAAD-INESC TEC, Porto, Portugal\\
         \email{cerqueira.vitormanuel@gmail.com}
         \and
         Lu\'{i}s Torgo \at
         Dalhousie University, Halifax, Canada\\
         LIAAD-INESC TEC, Porto, Portugal\\
         University of Porto, Porto, Portugal \\
         \email{ltorgo@dcc.fc.up.pt}
         \and
         Carlos Soares \at
         Fraunhofer AICOS Portugal, Porto, Portugal\\
         INESC TEC, Porto, Portugal\\
         University of Porto, Porto, Portugal\\
         \email{csoares@fe.up.pt}
         }

\date{Received: date / Accepted: date}

\maketitle

\begin{abstract}

The early detection of anomalous events in time series data is essential in many domains of application. 
In this paper we deal with critical health events, which represent a significant cause of mortality in intensive care units of hospitals. 
The timely prediction of these events is crucial for mitigating their consequences and improving healthcare.
One of the most common approaches to tackle early anomaly detection problems is standard classification methods. 
In this paper we propose a novel method that uses a layered learning architecture to address these tasks. One key contribution of our work is the idea of pre-conditional events, which denote arbitrary but computable relaxed versions of the event of interest. We leverage this idea to break the original problem into two hierarchical layers, which we hypothesize are easier to solve. 
The results suggest that the proposed approach leads to a better performance relative to state of the art approaches for critical health episode prediction.

\end{abstract}

\section{Introduction}

\subsection{Motivation for Early Anomaly Detection in Healthcare}

Healthcare is one of the domains which has witnessed significant growth in the application of machine learning approaches \cite{bellazzi2008predictive}. For instance, ICUs evolved considerably in recent years due to technological advances such as the widespread adoption of bio-sensors \cite{saeed2002mimic}. This lead to new opportunities for predictive modelling in clinical medicine. One of these opportunities is the early detection of critical health episodes (CHE), such as acute hypotensive episode \cite{ghosh2016hypotension} (AHE) or tachycardia episode \cite{forkan2017visibid} (TE) prediction problems. CHEs such as these represent a significant mortality risk factors in ICUs \cite{ghosh2016hypotension}, and their timely anticipation is fundamental for improving healthcare.

AHE or TE prediction can be regarded as a particular instance of early anomaly detection in time series data, also known as activity monitoring \cite{fawcett1999activity}. The goal behind these problems is to issue accurate and timely alarms about interesting but rare events requiring action. In the case of CHE, a system should signal physicians about any impending health crisis.

One of the most common ways to address activity monitoring problems is to view them as conditional probability estimation problems \cite{fawcett1999activity,tsurhypotensive}. Standard supervised learning classification methods can be used to tackle them. The idea is to approximate a function $f$ that maps a set of input observations $X$ to a binary variable $y$, which represents whether an anomaly occurs or not. In the context of CHE prediction, the predictor variables ($X$) summarise the recent physiological signals of a patient assigned to the ICU, while the target ($y$) represents whether or not there is an impending event in the near future.

\subsection{Working Hypothesis and Approach}

\begin{figure}[!hbt]
    \centering
    \includegraphics[width=.55\textwidth, trim=0cm 0cm 0cm 0cm, clip=TRUE]{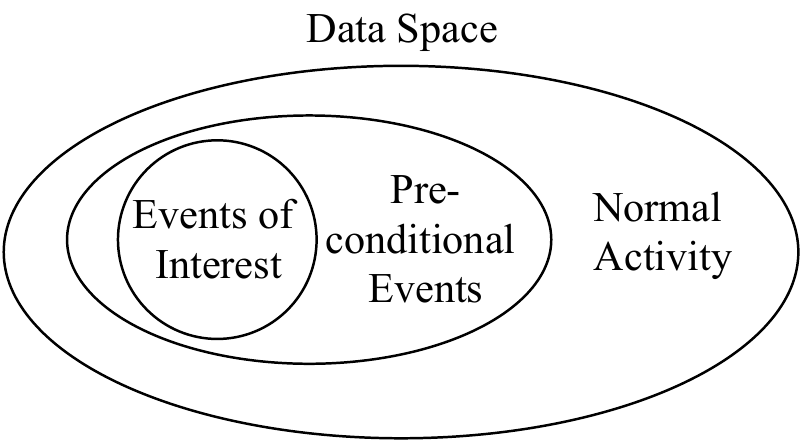}
    \caption{Essential scheme of the proposed layered learning approach. Instead of directly modelling events of interest according to normal activity, we first model pre-conditional events which co-occur with the events of interest and are more frequent.} 
    \label{fig:introscheme}
\end{figure}

In many domains of application, the anomaly or event of interest is defined according to some rule derived from the data by professionals. In the case of healthcare, CHEs are often defined as events where the value of some physiological signal exceeds a pre-defined threshold for a prolonged period. 
Similar approaches for formalising anomalies can be found in predictive maintenance \cite{ribeiro2016sequential}, or wind power prediction \cite{ferreira2011survey}. In these scenarios, we can also define pre-conditional events, which are arbitrary but computable relaxed versions of the event of interest. 
These pre-conditional events co-occur with the anomaly one is trying to model, but are more frequent and, in principle, a good indication for these. To be more precise, a pre-conditional event (i) represents a less extreme version of the anomalies we are trying to detect (main events); and (ii) co-occur with anomalies (i.e. there can not be an anomaly without a pre-conditional event). This concept is illustrated in Figure \ref{fig:introscheme} as a Venn diagram for classes.

Our working hypothesis in this paper is that modelling these pre-conditional events can be advantageous to capture the actual events of interest. To achieve this, we adopt a layered learning method \cite{stone2000layered}. 
Layered learning denotes a hierarchical machine learning approach in which a predictive task is split into two or more layers (simpler predictive tasks) where the learning process within a layer affects the learning process of the next layer. 
This type of approach is also common in the hierarchical reinforcement learning literature \cite{dietterich2000hierarchical}; for example, the options framework by Sutton \cite{sutton1998between}. Our contribution is its application to early anomaly detection problems.

Our approach exploits the idea that rare events of interest co-occur with pre-conditional events, which are considerably more frequent. Further, the same type of event of interest can be caused by distinct factors. For example, a particular type of CHE affecting two people may be caused by different diseases, which in turn may cause distinct dynamics in the time series of physiological signals. Therefore, initially modelling a relaxed version of the event of interest may lead to a simplification of the predictive task and a better performance when capturing the actual event of interest.

We apply the proposed approach to tackle the problem of CHE prediction. Our results show that, when comparing to the typical direct classification approach (without layered learning), the layered learning model leads to a significantly better event recall (more CHEs are timely predicted), with a comparable false alarm rate. The proposed method also shows an overall better predictive performance relative to other state of the art methods.

Typically, model for predicting CHEs are evaluated using standard classification metrics. In this work, we make a case concerning with the inappropriateness of these measures for evaluating early event detection model in high frequency time series. Alternatively, we adopt event recall and reduced precision \cite{weiss1998learning} to this effect. 

In short, the contributions of this paper are the following:
\begin{itemize}
    \item a general hierarchical approach to the early detection of anomalies in time series data;
    \item the application of the proposed approach to AHE and TE prediction;
    \item a discussion regarding the evaluation of activity monitoring methods in high frequency time series;
    \item a set of experiments validating the approach, including a comparison with state of the art approaches, a scalability analysis in terms of run-time, and a study of the impact of re-sampling strategies.
\end{itemize}

This paper is structured as follows. In the next section, we start by formalising the problem of activity monitoring, both in general terms and using the case study of event prediction in ICUs. In Section \ref{chp6:sec:ll}, we present the proposed layered learning approach to activity monitoring. We overview layered learning as proposed by Stone and Veloso \cite{stone2000layered}, and formalise our proposed adaptation for the early detection of CHE. In Section \ref{chp6:sec:experiments}, we carry out some experiments using the MIMIC II database \cite{saeed2002mimic}. In Section \ref{chp6:sec:relatedwork}, we overview the related work, and finally conclude the paper in Section \ref{chp6:sec:conclusions}.

We note that this paper is an extension of the work published in \cite{cerqueira2019layered}.
The data is publicly available by \cite{saeed2002mimic}. We also publish our code in an online repository\footnote{At \url{https://github.com/vcerqueira/layered_learning_time_series}}.

\section{Early Anomaly Detection}\label{chp6:sec:earlyanomaly}

We formalise the problem of early anomaly detection in time series in this section. We start by formalising the general problem and then the particular instances of AHE and TE prediction.

We follow Weiss and Hirsh \cite{weiss1998learning} to formalise the predictive task. 
Let $\mathcal{D} = \{D_1, \dots,$ $ D_{|\mathcal{D}|}\}$ denote a set of time series. In our case, $\mathcal{D}$ represents a set of patients being monitored at the ICU of an hospital.
Each $D_i \in \mathcal{D}$ denotes a time series $D_i = \{d_{i,1}, d_{i,2}, d_{i,n_i}\}$, where $n_i$ represents the number of observations for entity $D_i$, and each $d \in D_i$ represents information regarding $D_i$ in the respective time step (e.g. a set of physiological signals captured from a patient in the ICU). 

Each $D_i$ can also be represented as a set of sub-sequences $D_i = \{\delta_1, \delta_2, \dots,$ $\delta_i, \dots, \delta_{n'-1}, \delta_{n'}\}$ , where $\delta_i$ represents the $i$-\textit{th} sub-sequence. 
A sub-sequence is a tuple $\delta_i = (t_i, X_i, y_i)$, where $t_i$ denotes the time stamp that marks the beginning of the sub-sequence, $X_i \in \mathbb{X}$ represents the input (predictor) variables, which summarise the recent past dynamics of the time series $D_i$; and $y_i \in \mathbb{Y}$ denotes the target variable, which is a binary value ($y_i \in \{0, 1\}, \forall$ $i \in \{1, \dots, n'\}$) that represents whether or not there is an impending anomaly in the near future in the respective time series. How near in the future is typically a domain-dependent parameter. 
For each sub-sequence $\delta_i$, we construct the feature-target pair ($X_i$,$y_i$) as follows.

\begin{figure}[!hbt]
    \centering
    \includegraphics[width=\textwidth, trim=0cm 0cm 0cm 0cm, clip=TRUE]{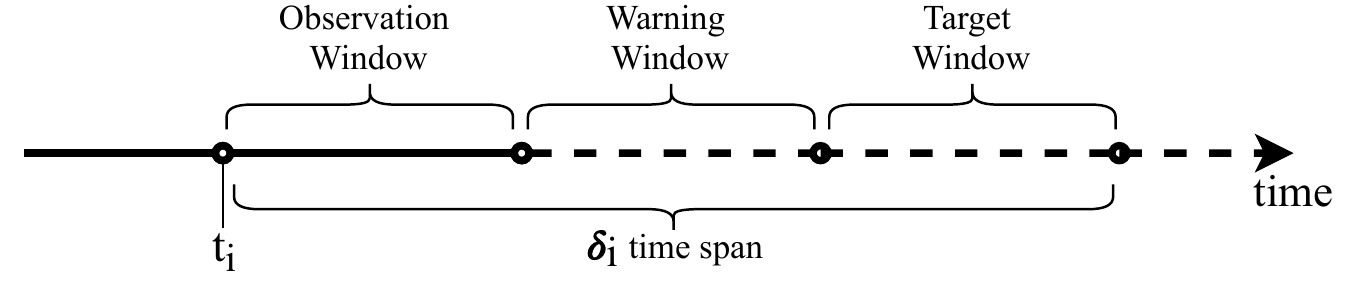}
    \caption{Splitting a sub-sequence $\delta_i$ into observation window, warning window, and target window. The features $X_i$ are computed during the observation window, while the outcome $y_i$ is determined in the target window. In the timeline, the solid line describes the data available (past), while the dotted line represents the future.} 
    \label{fig:windows}
\end{figure}

As illustrated in Figure \ref{fig:windows}, $\delta_i$ has three associated windows: (i) the target window (TW), which is used to determine the value of $y_i$; (ii) an observation window (OW), which is the period available for computing the values of $X_i$; and (iii) a warning window (WW), which is the lead time necessary for a prediction to be useful. An adequate WW enables a more efficient allocation of resources. Further, in the case of clinical medicine, physicians need some time after an alarm is launched to decide the most appropriate treatment.

The sizes of these windows depend on the domain of application and on the sampling frequency of the time series. In principle, the problem will be easier as the observation window is closer to the target window, that is, a smaller warning window is required. Weiss and Hirsh \cite{weiss1998learning} provide evidence for this property when predicting equipment failure, and Lee and Mark \cite{lee2010investigation} obtain similar results regarding AHE prediction. 

\subsection{Event Prediction in ICUs}\label{chp6:sec:ahe_pd}

In this work, we focus on a particular instance of early anomaly detection problems: CHE prediction in ICUs, namely AHE and TE. Ghosh et al. \cite{ghosh2016hypotension} state that prolonged hypotension leads to critical health damage, from cellular dysfunction to severe injuries in multiple organs. In turn, sustained tachycardia significantly increases the risk of stroke or cardiac arrest. Because CHEs are a relevant cause of mortality in ICUs, it is fundamental to anticipate them early in time so that physicians can prevent them or mitigate their consequences.

\begin{figure}[!hbt]
    \centering
    \includegraphics[width=\textwidth, trim=0cm 0cm 0cm 0cm, clip=TRUE]{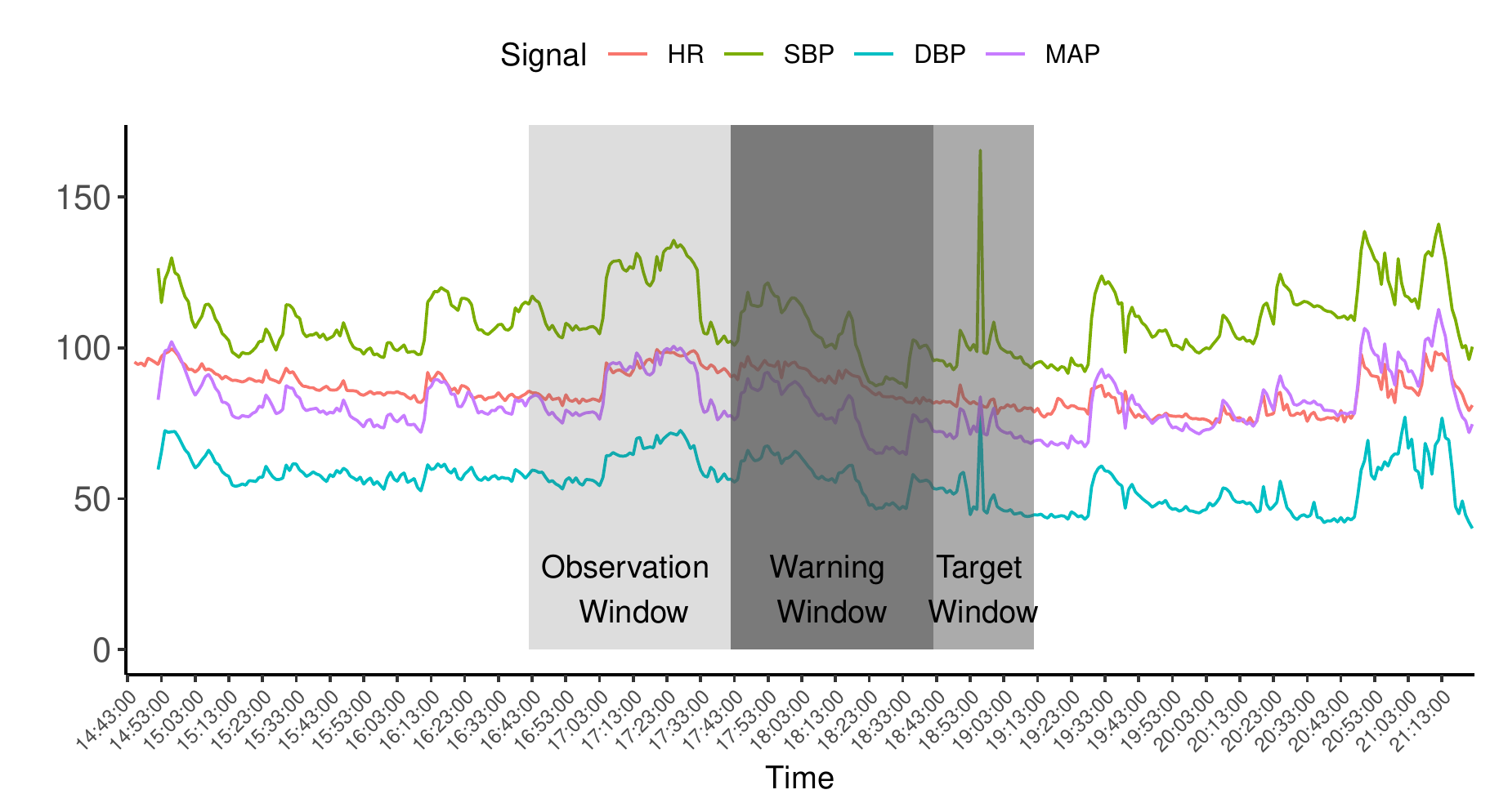}
    \caption{The physiological signal of patients are monitored over time. Each sub-sequence, denoted by the shaded areas, is split in an observation window, a warning window, and a target window.}
    \label{fig:eventsequence}
\end{figure}

Patients assigned to the ICU are typically continuously monitored, with bio-sensors capturing several physiological signals, such as heart rate, or mean arterial blood pressure.
This is illustrated in Figure \ref{fig:eventsequence}, where the data of a patient is depicted. A sub-sequence for CHE prediction is given as an example in the shaded area of the graphic. This area is split into three windows (observation, warning, target), as explained above. 

Sensors capturing the physiological signals of a patient are typically collected with a high sampling frequency. In this work, we assume a minute by minute sampling frequency.

\subsubsection{Acute Hypotensive Episodes}\label{chp6:sec:ahedef}

Hypotension episodes are defined as a prolonged drop in the blood pressure. More formally, AHE is an event defined as ``a 30-minute window having at least 90\% of its mean arterial blood pressure (MAP) values below 60 mmHg [millimetres of mercury]'' \cite{tsurhypotensive,lee2010investigation}. In this context, the target variable value is computed as follows:

\begin{equation*}
    y_i = \begin{cases}
            1, & \text{if an AHE happens in } TW_i,\\
            0, & \text{otherwise}.
          \end{cases}
\end{equation*}

\noindent In other words, we consider that the \textit{i}-th sub-sequence represents an anomaly if its target window represents an AHE (c.f. Figure \ref{fig:eventsequence}). Since AHEs are rare, the target vector $y$ is dominated by the negative class (i.e. $y = 0$), where a patient shows a normotensive status. For the target window of 30 minutes, we consider an observation window and a warning window of 60 minutes each. These values are typically used in the literature of AHE prediction models \cite{ghosh2016hypotension}. 

\subsubsection{Tachycardia Episodes}\label{sec:tedef}

Tachycardia denotes a high heart rate (HR). Generally, an HR over 100 beats per minute (bpm) under a resting state is considered tachycardia. In order to consider a more robust definition for the purpose of discovering tachycardia episodes, we follow a similar intuition to AHEs. We define TE as ``a 30-minute window having at least 90\% of its HR values above 100 bpm". The respective target variable is computed as follows:

\begin{equation*}
    y_i = \begin{cases}
            1, & \text{if an TE happens in } TW_i,\\
            0, & \text{otherwise}.
          \end{cases}
\end{equation*}

\noindent TEs are defined similarly to AHEs. Moreover, TEs also denote rare events since ICU patients usually show an HR below 100 bpm. We consider identical window sizes (OW, WW, TW) for both problems.

\subsection{Discriminating Approaches to Early Anomaly Detection}\label{sec:methods}

Naturally, one of the most common approaches to solving the problem defined previously is to view it as a conditional probability estimation problem and use standard supervised learning classification methods \cite{fawcett1999activity,tsurhypotensive}. The idea is to build a model $f: \mathbb{X} \rightarrow \mathbb{Y}$, where $X \in \mathbb{X}$ and $y \in \mathbb{Y}$. This model can be used to predict the target values associated with unseen feature attributes. In other words, $f$ is a discriminating model that explicitly distinguishes normal activity from anomalous activity.

Notwithstanding the widespread of this approach, early anomaly detection problems often comprise complex target variables whose definition is derived from the data. In such cases, it is possible to decompose the target variable into partial and less complex concepts, which may be easier to model. In this context, our working hypothesis is that we can leverage a layered learning approach to model these partial concepts and obtain an overall better model for capturing the actual events of interest.

\section{Layered Learning for Early Anomaly Detection}\label{chp6:sec:ll}

\subsection{Layered Learning}

Layered learning is designed for predictive tasks whose mapping from inputs to outputs is complex. For example, Stone and Veloso \cite{stone2000layered} apply this approach to robotic soccer. Particularly, one of the problems they face is the retrieval and passing of a ball. The authors split this task into three layers: (i) ball interception; (ii) pass evaluation; and (iii) pass selection. This process leads to a more effective decision-making system with a considerably higher success rate than a direct approach. 
This type of approach is also common in the reinforcement learning literature \cite{dietterich2000hierarchical}. Specifically, hierarchical reinforcement learning consists in breaking a problem into a hierarchy of sub-tasks.

As Stone and Veloso \cite{stone2000layered} describe, ``the key defining characteristic of layered learning is that each layer directly affects the learning of the next''. This effect can occur in several ways; for example, by affecting the set of training examples, or by providing features used for learning the original concept.

The general assumption behind decomposing a problem into hierarchical sub-tasks is that the problem addressed in each layer is simpler than the original one. We hypothesise that, when combining the models in each layer, this leads to a better overall approach for solving the task at hand.

\subsection{Pre-conditional Events}

The definition of an anomalous event in time series data is in many cases determined according to some rule derived from the data. As an example from the healthcare domain presented in the previous section, an AHE is defined as a percentage of numeric values which are below some threshold within a time interval (c.f. Section \ref{chp6:sec:ahedef}). TEs are defined similarly. This type of approach for defining anomalous events is also common in other domains. For example, in predictive maintenance \cite{ribeiro2016sequential}, where numerical information from sensor readings is transformed into a class label which denotes whether or not an observation is anomalous. Or wind ramp detection, where a ramp event is a rare occurrence that denotes a large percentage change in wind power output in a short time interval \cite{ferreira2011survey}.

Since these anomalous events are defined according to the value of an underlying variable, we can also define pre-conditional events: relaxed versions of the actual events of interest, but which are more frequent. A more precise definition can be given as follows. A pre-conditional event is an arbitrary but computable event that is expected to occur with the main event taking place simultaneously. If the main event occurs, the pre-conditional event must occur, but the latter can occur without the main event.

An example can be provided using the case study of AHE prediction. In Section \ref{chp6:sec:ahedef}, we defined the main event (AHE) as ``a 30-minute window having at least 90\% of its mean arterial blood pressure (MAP) values below 60 mmHg''. A possible pre-conditional event for this scenario could be ``a 30-minute window having at least \textbf{45}\% of its mean arterial blood pressure (MAP) values below 60 mmHg''. Another possibility is ``a 30-minute window having at least 90\% of its mean arterial blood pressure (MAP) values below \textbf{70} mmHg''.

In summary, pre-conditional events should have the following two characteristics:
\begin{itemize}
    \item pre-conditional events should have a higher relative frequency than the main events;
    
    \item pre-conditional events always happen when the main events happen. The inverse is not a necessary condition.
\end{itemize}

\subsection{Methodology}

We can leverage the idea of pre-conditional events and use a layered learning strategy to tackle activity monitoring problems in time series data. Our idea is to decompose the main predictive task into two layers, each denoting a predictive sub-task. Pre-conditional events are modelled in the first layer, while the main events are modelled in the subsequent one.

\begin{figure}[!hbt]
    \centering
    \includegraphics[width=\textwidth, trim=0cm 0cm 0cm 0cm, clip=TRUE]{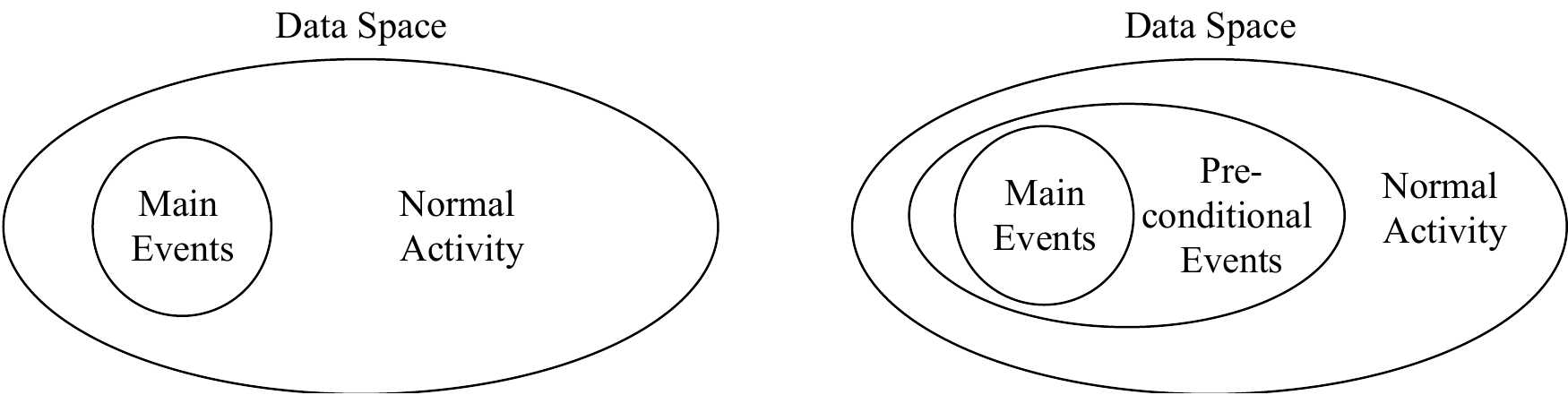}
    \caption{Venn diagram for the classes in an activity monitoring problem. The main event represents a small part of the data space; pre-conditional events are more frequent and include the occurrence of the main events.}
    \label{fig:LLDATASPACE}
\end{figure}

The intuition behind this idea is given in Figure \ref{fig:LLDATASPACE}. The figure presents two Venn diagrams for classes. Focusing on the left-hand side, the anomalies or main events (e.g. AHE) represent a small part of the data space. This is one of the issues that makes them difficult to model. In the typical classification approach, main events are directly modelled with respect to the remaining data space (deemed normal activity). 

Our idea is represented on the right-hand side. An initial pre-conditional concept is considered, which is more common than the main target concept, while also including it. The higher relative frequency of the pre-conditional events with respect to the main events helps to mitigate the problem of having an imbalanced distribution, which is the case in activity monitoring tasks. This phenomenon can compromise the performance of learning algorithms \cite{he2013imbalanced}. In effect, we first model the pre-conditional events with respect to normal activity. These pre-conditional events are, in principle, easier to learn relative to the main concept because they are more frequent and thus the classification algorithms will not suffer so much from an imbalanced distribution. Afterwards, the main target events are modelled with respect to the pre-conditional events, which is also a less imbalanced distribution than the original on the left diagram.

In the remainder of this section, we will further formalise our approach using a generic notion of pre-conditional and main events. In the next section, we will apply this formalisation to CHE prediction problems.

\subsubsection{Pre-conditional Events Sub-task}\label{chp6:sec:prec_task}

Let $\mathcal{S}$ denote a pre-conditional event. The target variable when modelling these events is defined as:

\begin{equation}\label{eq:task1b}
  y^{\mathcal{S}}_i = \begin{cases}
            1 & if \text{ } \mathcal{S} \text{ happens},\\
            0 & \text{otherwise}.
          \end{cases}
\end{equation}

For this task, a sub-sequence $\delta^\mathcal{S}_i$ is a tuple $\delta^\mathcal{S}_i = (t_i, X_i, y^\mathcal{S}_i)$. The difference to the original sub-sequence $\delta_i$ is the target variable, which replaces $y$ with $y^\mathcal{S}$. Finally, the goal of this first predictive task is to build a function $f^\mathcal{S}$ that maps the input predictors $X$ to the output $y^\mathcal{S}$.

\subsubsection{Main Events Sub-task}

Provided that we solve the pre-conditional events sub-task, in order to predict impending main events the remaining problem is to find out whether or not, when $\mathcal{S}$ happens, the main event also happens. 

Let $\mathcal{F}$ be defined as the occurrence: ``given $\mathcal{S}$, there is an impending main event in the target window of the current sub-sequence''. Effectively, the target variable for this task is defined as follows:

\begin{equation}\label{eq:task2}
 \text{Given }y^{\mathcal{S}}\text{ = 1, } y^{\mathcal{F}}_{i} = \begin{cases}
            1 & \text{if a main event happens in } TW_i,\\
            0 & \text{otherwise}.
          \end{cases}
\end{equation}

The target variable for this sub-task ($y^{\mathcal{F}}$) is formalised in equation \ref{eq:task2}. Given that the class of $y^{\mathcal{S}}$ is positive (which means that there is an impending pre-conditional event), the class of $y^{\mathcal{F}}$ is positive if a main event also happens in that same target window, or negative otherwise.

The goal of this second predictive task is to build a function $f^\mathcal{F}$, which maps $X$ to $y^\mathcal{F}$. 
Formally, a sub-sequence $\delta^\mathcal{F}_i$ is represented by $\delta^\mathcal{F}_i = (t_i, X_i, y^\mathcal{F}_i)$. In this scenario, however, the set of available sub-sequences $D_i$ is considerably less than in the pre-conditional sub-task because only the sequences for which $y^\mathcal{S}$ equals 1 are accounted for. 
This means that $f^\mathcal{F}$ only learns with sub-sequences that at least lead to a pre-conditional event.
Effectively, this aspect represents how the learning in the pre-conditional events sub-task affects the learning on the main events sub-task, i.e., by influencing the data examples used for training. In the main events sub-task, a predictive model is concerned with the distinction between pre-conditional events and main events. Essentially, it assumes that the distinction between normal activity and pre-conditional events is carried out by the previous layer. Given this independence, the training of the two layers can occur in parallel.

\subsubsection{Predicting Impending Anomalies}

To make predictions about impending events of interest we combine the models $f^\mathcal{S}$ with $f^\mathcal{F}$ with a function $ g: \mathbb{X} \times \mathbb{X} \rightarrow \mathbb{Y}$.

\begin{equation}\label{eq:merget1t2}
    g(X_i) = f^\mathcal{S}(X_i) \cdot f^\mathcal{F}(X_i)
\end{equation}

Essentially, according to equation \ref{eq:merget1t2} the function $g$ predicts that there is an impending main event in a given sub-sequence $\delta_i$ according to the multiplication of the outcome predicted by both $f^\mathcal{S}$ and $f^\mathcal{F}$. 

Ideally, there are three possible outcomes:
\begin{itemize}
    \item Both event $\mathcal{S}$ and event $\mathcal{F}$ happen, which means there is an impending main event: both $f^\mathcal{S}$ and $f^\mathcal{F}$ should return $1$ so that $f^\mathcal{S} \cdot f^\mathcal{F} = 1$;
    \item Event $\mathcal{S}$ happens, but event $\mathcal{F}$ does not happen: $f^\mathcal{S} = 1$, but $f^\mathcal{F} = 0$, so $f^\mathcal{S} \cdot f^\mathcal{F} = 0$;
    \item Event $\mathcal{S}$ does not happen, and consequently, event $\mathcal{F}$ also does not happen: $f^\mathcal{S} \cdot f^\mathcal{F} = 0$.
\end{itemize}

\subsection{Application of Layered Learning to CHE Prediction}

We formalise the application of our idea to CHE problems, namely AHE and TE prediction. 

\subsubsection{AHE Prediction}

As mentioned before (c.f. Section \ref{chp6:sec:ahedef}), an AHE is defined as a 30-min period where 90\% of the blood pressure values are below 60 mmHg. We propose to relax this threshold and define the pre-conditional event $\mathcal{S}$ as:
\begin{description}
    \item  $\mathcal{S}^{\text{AHE}}$: ``a 30-minute window having at least \textbf{45}\% of its mean arterial blood pressure values below 60 mmHg". 
\end{description}

The event $\mathcal{S}$ is consistent with the two above-mentioned characteristics: the frequency of $\mathcal{S}$ across the database is considerably higher than an AHE -- note that the blood pressure level can drop below 60 mmHg for some time period without being considered as a hypotensive episode. Consequently, the occurrence $\mathcal{S}$ is simultaneous to the occurrence of an AHE (if 90\% of the values are below 60 mmHg, so are 45\%). 

\subsubsection{TE Prediction}

We apply the same reasoning to the TE prediction task. In Section \ref{sec:tedef}, we defined a TE as ``a 30-minute window having at least 90\% of its HR values above 100 bpm". In order to define $\mathcal{S}^{\text{TE}}$ we again relax the percentage threshold as follows:
\begin{description}
    \item  $\mathcal{S}^{\text{TE}}$: ``a 30-minute window having at least 45\% of its HR values above 100 bpm". 
\end{description}

\noindent Similarly to $\mathcal{S}^{\text{AHE}}$, the events $\mathcal{S}^{\text{TE}}$ also follow the desired characteristics of pre-conditional events. 

In both situations (AHE and TE), the value of 45\% was chosen arbitrarily. 
Essentially, we attempted to make the pre-conditional events much more frequent relative to the main events.
Nevertheless, this parameter can be optimised.
Similarly to other hierarchical methodologies, for example in hierarchical reinforcement learning \cite{dietterich2000ensemble}, the definition of the sub-task is performed manually. In Section \ref{sec:towards}, we will discuss this issue further.

\section{Empirical Experiments}\label{chp6:sec:experiments}

\subsection{Case Study: MIMIC II}

In the experiments, we used the database Multi-parameter Intelligent Monitoring for Intensive Care (MIMIC) II \cite{saeed2002mimic}, which is a benchmark for several predictive tasks in healthcare, including CHE prediction. 

As inclusion criteria of patients and general database pre-processing steps, we follow Lee and Mark \cite{lee2010investigation} closely. For example, the sampling frequency of the physiological data of each patient in the database is minute by minute. Moreover, the following physiological signals are collected: heart rate (HR), systolic blood pressure (SBP), diastolic blood pressure (DBP), and mean arterial blood pressure (MAP).

As described in Section \ref{chp6:sec:ahe_pd}, the target window size is 30 minutes. For each target window, there is a 60-minute observation window and a 60-minute warning window. For a comprehensive read regarding the data compilation, we refer to the work by Lee and Mark \cite{lee2010investigation}. Considering this setup, the number of patients is 1,072, leading to a data size of 1,975,936 sub-sequences. 71,035 of those sub-sequences represent an AHE (about 3.5\%). In turn, 13.6\% of the sub-sequences represent a TE.


We consider HR, SBP, DBP, and MAP values between 10 and 200 (bpm for HR, mmHg for the remaining ones). Values outside of this range are eliminated as ``unlikely outliers'' \cite{lee2010investigation}. From the available signals (HR, SBP, DBP, MAP), we compute the values of cardiac output (CO) and pulse pressure (PP). 

Regarding feature engineering, we follow previous work in the literature \cite{lee2010hypotensive,tsurhypotensive}. Using the observation window of each sub-sequence and of each physiological signal, the feature engineering process was carried out using statistical, cross-correlation, and wavelet functions. The statistical metrics include skewness, kurtosis, slope, median, minimum, maximum, variance, mean, standard deviation, and inter-quartile range. For each observation window, we also compute the cross-correlation of each pair of signals at lag 0. We also use the Daubechies wavelet transform \cite{percival2006wavelet} to perform a 5-level discrete wavelet decomposition and capture the relative energies in different spectral bands. Intuitively, medication data can play an important role. However, Lee and Mark \cite{lee2010hypotensive} reported no predictive advantage in using such information. In effect, we do not include this information in the predictive models. 

\subsection{Experimental Design}

The experiments were designed to answer the following research questions:
\begin{description}
    \item[\textbf{RQ1}:] how does the proposed layered learning architecture performs relative to state of the art approaches for early anomaly detection?
    \item[\textbf{RQ2}:] what is the predictive performance of each layer in the proposed layered learning architecture for CHE prediction?
    \item[\textbf{RQ3}:] how does the layered learning approach scale in terms of run-time compared to other approaches?
    \item[\textbf{RQ4}:] what is the impact of pre-processing the training data using a re-sampling method for balancing the distribution of classes?
\end{description}

To estimate the predictive performance of each method, we used a 5$\times$10-fold cross-validation, in which folds are split by patients. To be more precise, in each iteration of the cross-validation procedure, one fold of patients is used for validation, another fold of different patients is used for testing, and the remaining patients are used for training the predictive model. Therefore, all sub-sequences of a given patient are only used for either training, validation, or testing. The set of time series (patients) only comprises a temporal dependency within each patient, and we assume the data across patients to be independent. 
That is, the probability that a patient suffers a health crisis is independent of another patient also suffering a health crisis. 
In this context, the application of cross-validation in this setting is valid. Finally, the sub-sequences of the patients chosen for training are concatenated together to fit the predictive model. This model is tuned using the sub-sequences of patients chosen for validation and evaluated using the sub-sequences of patients chosen for testing. 

\subsubsection{Sub-sequences used for Training}\label{sec:ss4tr}

Given the sizes of OW, WW, and TW (60, 60, and 30 minutes, respectively), the duration of a sub-sequence is 150 minutes. Since the data is collected every minute, there is considerable overlap between consecutive sub-sequences. During run-time, a given model is used to predict whether there is an impending CHE in each sub-sequence. This approach emulates a realistic scenario, where a prediction is produced as more data is available regarding the current health state of a given patient.

Given the redundancy among consecutive sub-sequences, it is common to sample the sub-sequences used for training a predictive model \cite{tsurhypotensive}. For example, Cao et al. \cite{cao2008predicting} compile sub-sequences for training according to whether a patient has experienced a CHE. For every patient that did, the latest 120 minutes of data before the onset of the respective CHE are used to create a training sub-sequence. If a patient did not experience a CHE, one or more sub-sequences are sampled at random. Lee and Mark \cite{lee2010hypotensive} collect multiple sub-sequences in a sliding window fashion, irrespective of whether a patient experienced a CHE. A sliding window with no overlap and of size TW is used to traverse each patient. That is, if a sub-sequence $\delta_i$ starts at time $t_j$ then the next sub-sequence $\delta_{i+1}$ starts at time $t_{j+30}$. The authors show that this approach leads to better results relative to the approach taken by Cao et al. \cite{cao2008predicting}.

In both cases described above, the authors note that these approaches lead to an imbalanced data set. They recommend under-sampling the majority class to overcome this issue. In this work, we follow the approach by Lee and Mark \cite{lee2010hypotensive}. As recommended, we also apply a class balance procedure, which is described below in Section \ref{sec:ll:sota} and analysed in Section \ref{sec:resampleanalysis}.

\subsubsection{The Value of a Prediction}

The timely prediction of impending CHEs enables a more efficient allocation of ICU resources and a more prompt application of the appropriate treatment. In this context, for a prediction to be useful, it must occur before the onset of the respective CHE. We assume that, after the event starts, any prediction becomes obsolete. Further, predicting too early also leads to meaningless predictions due to the continuity of time. We follow the approach taken in the $10^{th}$ PhysioNet challenge \cite{moody2009predicting} regarding AHE prediction. A CHE (we also extend the definition to TEs) is considered to be correctly anticipated if it starts within 60 minutes after an alarm is launched. We consider the value of an alarm to be binary, where its benefit is 1 if it is issued correctly, and 0 otherwise.

\subsubsection{Learning Algorithms}

We tested different predictive models in the experiments, namely a random forest \cite{ranger2015}, a support vector machine \cite{kernlab04}, a deep feed-forward neural network \cite{abadi2016tensorflow}, and an extreme gradient boosting (\texttt{xgboost}) model \cite{chen2015xgboost}. We only show the results of the latter in these experiments, since it provides better performance than the remaining methods for both AHE prediction and TE prediction. This corroborates the experiments by Tsur et al. \cite{tsurhypotensive}, stating that \texttt{xgboost} gives the best predictive performance for AHE prediction tasks.

The output of the classifiers used in the experiments is a probability.
The decision threshold is optimised following previous works in the literature of AHE prediction \cite{lee2010investigation,tsurhypotensive}, which recommend selecting the threshold that maximises the average of recall and specificity (true negative rate).

\subsection{State of the Art Methods}\label{sec:ll:sota}

We compare the proposed layered learning approach (henceforth denoted as \texttt{LL}) with the following four methods.

\subsubsection{Standard classification}

We compare \texttt{LL} with a standard classification method (\texttt{CL}) that does not apply a layered learning approach and directly models the events of interest with respect to normal activity (c.f. Figure \ref{fig:LLDATASPACE}). One of the working hypothesis for the application of the proposed layered learning approach is that it helps to mitigate the class imbalance problem. To further cope with this problem, we process the data used for training \texttt{CL} and \texttt{LL} using a re-sampling method \cite{ublpackage}. In the case of \texttt{LL}, this process was applied to both layers after performing the task decomposition. For the AHE prediction problem, \texttt{CL} was applied with random over-sampling, while \texttt{LL} was applied using random under-sampling. For TE prediction, both approaches were applied using SMOTE \cite{chawla2002smote}. These choices are analysed in Section \ref{sec:resampleanalysis}.

\subsubsection{Isolation Forest} 

An Isolation Forest (\texttt{IF}) \cite{liu2012isolation} is a state of the art unsupervised model-based approach to anomaly detection. A typical method of this sort typically discards the anomalies within the training data and creates a model for normal activity. Observations that significantly deviate from the typical behaviour are considered outliers. We referred to these approaches as profiling methods (Section \ref{sec:am}). Instead of separating the normal activity, \texttt{IF} explicitly models the anomalies in an unsupervised manner using an ensemble of tree-structured models. The core idea behind a \texttt{IF} is that the paths resulting from partitioning the data are shorter for anomalous observations because the regions comprising these anomalies are separated quickly.

\subsubsection{Regression Approach}

The type of anomalies addressed in this paper is defined according to the observed values in numeric variables. For example, an AHE is defined according to the distribution of the variable MAP in the target window of an sub-sequence. In effect, one common alternative modelling approach is to perform a regression analysis with the respective numeric variable as the target variable. Alarms regarding impending anomalies are then triggered using a deterministic function that maps the predicted values into a decision.

We include a regression-based alternative both for AHE prediction and TE prediction. We apply a multi-step forecasting model to predict the future values of MAP (for AHEs) and HR (for TEs) for the next TW. Regarding the former, and following up on the definition of an AHE (Section \ref{chp6:sec:ahedef}), an alarm for an AHE is triggered if 90\% of the predicted values for the MAP variable are below 60 mmHg \cite{rocha2011prediction}. Likewise, an alarm for a TE is triggered if 90\% of the predicted values for the HR variable are above 100 bpm. 

Similarly to Rocha et al. \cite{rocha2011prediction}, the multi-step forecasting model follows a \textit{direct} approach \cite{taieb2012review}. This means that a forecasting model is created for each point in the horizon. The horizon in our setup is represented by the target window, which is a 30-minute window of observations with a granularity by the minute. In effect, the regression-based approach is comprised of 30 forecasting models. We denote the regression-based model as \texttt{RG}. To train each forecasting model, we also use a \texttt{xgboost} learning algorithm, which is tuned for numeric target variables.

\subsubsection{Ad-hoc Methods}

While there is an increasing number of machine learning applications in healthcare, many of the currently deployed systems still rely on simple \textit{ad-hoc} rules to support the decision-making process of professionals. Taking AHE prediction as an example, a simple rule is to trigger an alarm if the MAP of a patient drops below 60 mmHg in a given time step. A similar approach can be used for TE prediction, where an alarm is launched if the HR variable exceeds 100 bpm. However simple, these ad-hoc rules often work well in practice. We use these rules as baselines in our experimental design and denote them as \texttt{AH}.

\subsection{Evaluation Metrics}

Approaches dealing with the prediction of critical health episodes typically evaluate predictive models using classical classification metrics, namely precision, recall, and F1 \cite{lee2010investigation,lee2010hypotensive,ghosh2016hypotension,tsurhypotensive}. However, these metrics are unsuitable when dealing with high frequency time series because, as Fawcett and Provost \cite{fawcett1999activity} state, they ignore the temporal order of observations and the value of timely decisions. We give particular examples which show cases where these metrics may report misleading results, and describe appropriate alternatives.

The goal behind early anomaly detection problems is not to classify each sub-sequence as positive or negative \cite{fawcett1999activity}. Instead, the main goal is to detect, in a timely manner, when there is an impending anomalous event. In this context, we follow Weiss and Hirsh \cite{weiss1998learning} regarding the evaluation metrics. Specifically, two measures are computed: Event Recall (ER), and Reduced Precision (RP). These two metrics follow the same intuition of the widely used Recall and Precision metrics but are tailored for time-dependent data.




    
    
      
      
      
      
      
      

\subsubsection{Event Recall}

Recall, also known as sensitivity, expresses how many positive cases are retrieved by a model. In our problem, a case represent a given sub-sequence. However, this metric may be misleading. Consider a scenario where model $m$ captures five positive different sub-sequences. The value of this model is different if these five predictions refer to the same event or to five different events. However, the recall metric, as usually applied, is blind to this nuance.

We adopt the event recall metric to overcome this problem. Let $T$ denote the total number of events of interest in a test data set, and let $\hat{T}_m$ represent the total number of those events correctly predicted by a model $m$. The ER for model $m$ is given by the following equation:
\begin{equation}
    \text{ER}_m = \frac{\hat{T}_m}{T}
\end{equation}

\noindent ER differs from the classical recall metric because a single correct prediction within an observation window leading to an event is enough to consider that event correctly anticipated. 
As Fawcett and Provost \cite{fawcett1999activity} put it, ``alarming earlier may be more beneficial, but after the first alarm, a second alarm on the same sequence may add no value''.

\subsubsection{Reduced Precision}

\begin{figure}[!hbt]
    \centering
    \includegraphics[width=.65\textwidth, trim=0cm 0cm 0cm 0cm, clip=TRUE]{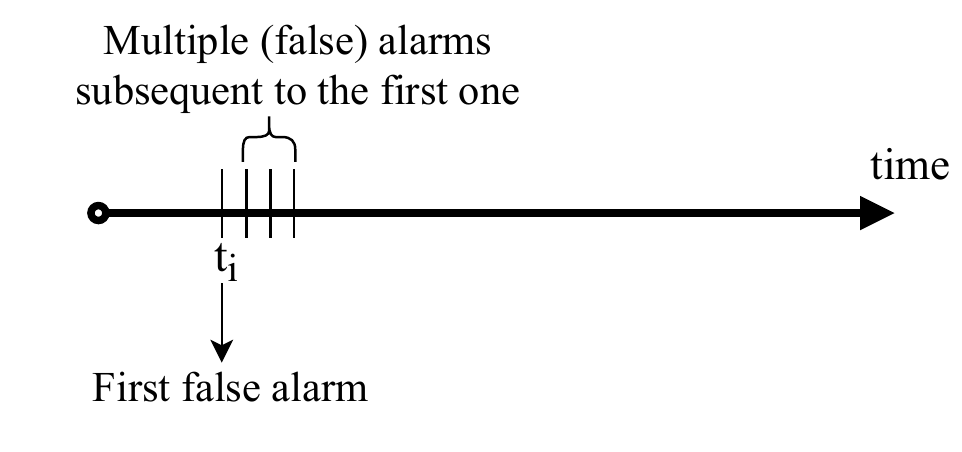}
    \caption{A sequence of consecutive false alarms. The first alarm is useful, but the subsequent ones may add no information.}
    \label{fig:rp_example}
\end{figure}

The classical precision metric measures the ratio of positive predictions that are correct. Similarly to recall, in a time-dependent domain, the classical precision may be misleading because multiple predictions on the same event are counted multiple times. This idea is intuited in Figure \ref{fig:rp_example}. This graphic shows a sequence in which predictions are being produced over time. Starting from time $t_i$, four false alarms are triggered. Performance evaluation should take the first wrong prediction into account as a false positive. However, the subsequent false alarms (as shown in Figure \ref{fig:rp_example}) are not meaningful since they add no information -- assuming some action is taken after the first alarm.

\begin{figure}[!hbt]
    \centering
    \includegraphics[width=.7\textwidth, trim=0cm 0cm 0cm 0cm, clip=TRUE]{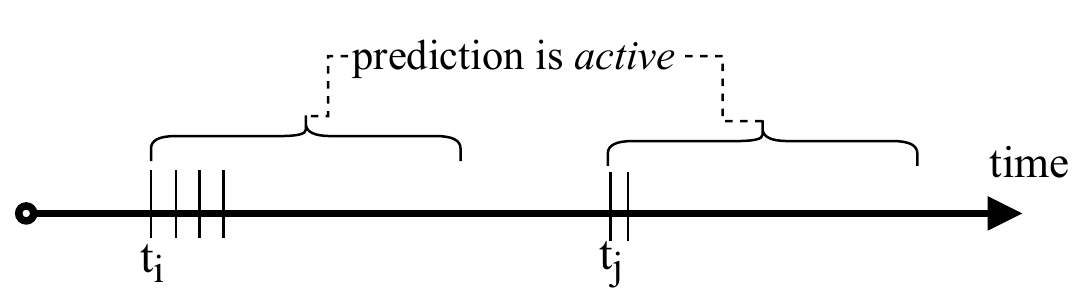}
    \caption{False alarms (denoted as vertical bars) over a time interval. There are 6 false positives, but only two discounted false positives.}
    \label{fig:rpactive}
\end{figure}

RP overcomes this problem by considering a prediction to be \textit{active} for some time period. Specifically, in this work, we consider a time interval with the same size as the observation window (60 minutes). Notwithstanding, this is usually a domain-dependent parameter. Effectively, the RP metric replaces the number of false positives with the number of discounted false positives -- the number of non-overlapping observation periods associated with a false prediction. This idea is illustrated in Figure \ref{fig:rpactive}, where each vertical bar in the time line denotes an issued false alarm. There are a total of 6 false positives, but, if taking into account the time interval a prediction is active, there are only two discounted false positives (DFP). Finally, RP also considers the number of target events correctly identified ($\hat{T}_m$), instead of the number of correct predictions (true positives).

\noindent In effect, RP for model $m$ is given by the following equation:

\begin{equation}
  \text{RP}_m = \frac{\hat{T}_m}{\hat{T}_m + \text{DFP}_m}
\end{equation}

ER and RP summarise the predictive performance. To further explore the behaviour of a given method, we also compute the average number of false alarms (FA); and the average anticipation time (AT) (how long in advance an event is predicted).

\subsection{Results on AHE Prediction}

Table \ref{tab:ahe_results} presents the average results, and respective standard deviation, for each method across the 50 folds ($5\times10-$fold CV) for the AHE prediction problem. 
Overall, \texttt{LL} presents the highest ER, capturing 83\% of the AHE. These values are significantly better relative to the remaining methods, including \texttt{CL}, which is the typical approach to solve these predictive tasks. 
Conversely, \texttt{LL} shows a comparable RP with \texttt{CL}. The RP shown by \texttt{LL} is better than \texttt{IF}'s, but considerably worse than the ones shown by \texttt{AH} and \texttt{RG}.

\begin{table}[!ht]
    \caption{Average of results for the AHE prediction problem across the 50 folds. Boldface represents the best result in the respective metric}
    \label{tab:ahe_results}
    \begin{tabular}{l|r|r|r|r}
      \textbf{Method} & \textbf{ER} & \textbf{RP} & \textbf{Avg. AT} & \textbf{Avg. FA}\\
      \hline
      
      \texttt{AH} & 0.625$\pm$0.057 & 0.129$\pm$0.022 & 31.9$\pm$3.3 & 4.9$\pm$2.2\\
      
      \texttt{CL} & 0.807$\pm$0.072 & 0.089$\pm$0.016 & 44.9$\pm$3.3 & 20.7$\pm$6.0\\
      
      \texttt{LL} & \textbf{0.830}$\pm$0.054 & 0.090$\pm$0.015 & \textbf{46.9}$\pm$3.3 & 22.9$\pm$6.9\\
      
      \texttt{RG} & 0.250$\pm$0.067 & \textbf{0.205}$\pm$0.044 & 11.1$\pm$3.2 & \textbf{3.3}$\pm$9.1\\
      
      \texttt{IF} & 0.700$\pm$0.182 & 0.035$\pm$0.009 & 37.9$\pm$11.8 & 26.6$\pm$15.2\\
    \end{tabular}%
\end{table}

\begin{figure}[!hbt]
    \centering
    \includegraphics[width=\textwidth, trim=0cm 0cm 0cm 0cm, clip=TRUE]{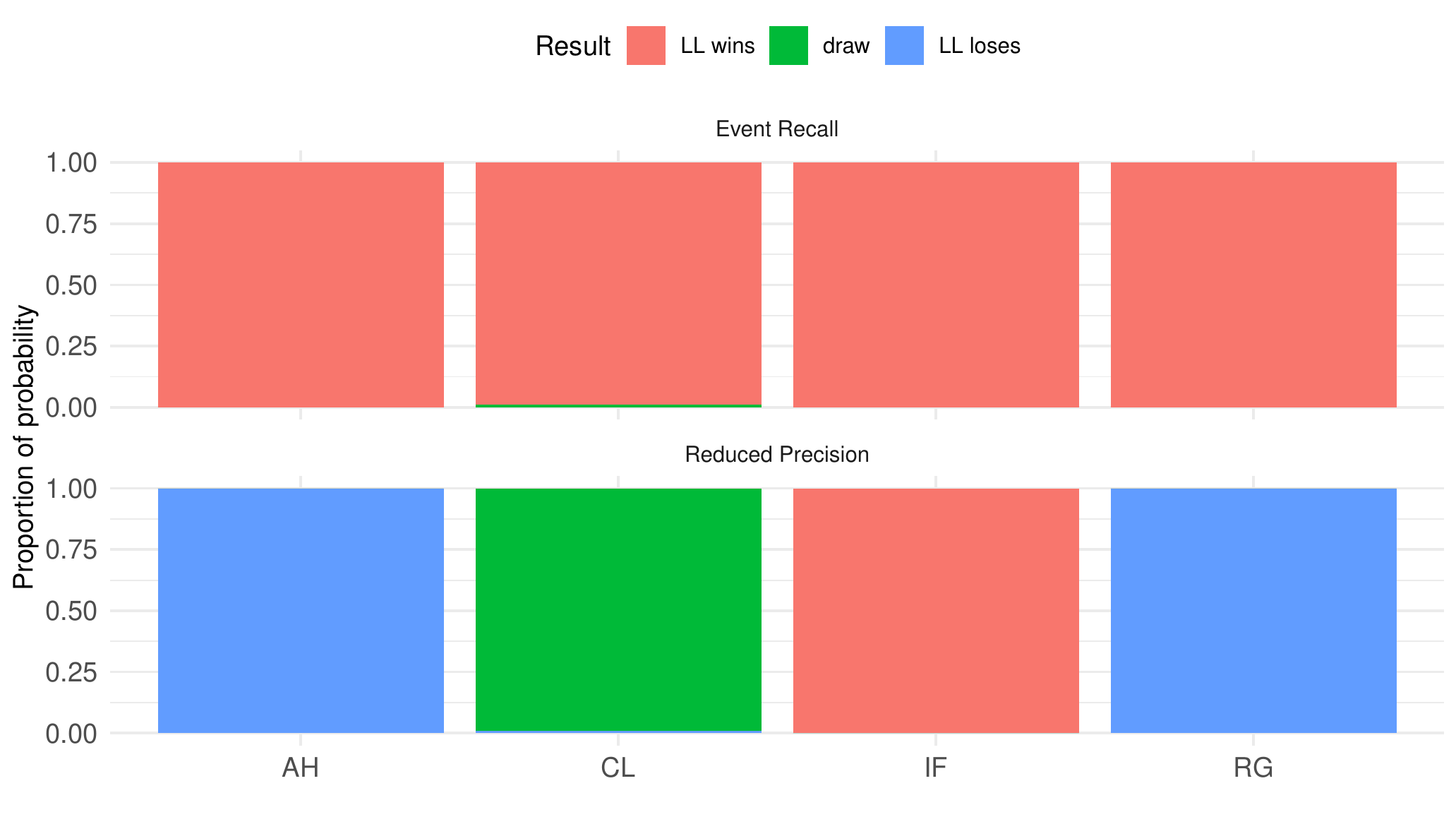}
    \caption{Comparing \texttt{CL} with \texttt{LL} with a Bayesian correlated t-test for ER and RP metrics (AHE prediction)}
    \label{fig:bayes_ahe}
\end{figure}

In Figure \ref{fig:bayes_ahe}, we analyse the significance of the results according to the Bayesian correlated t-test \cite{benavoli2017time}. In this test, we consider the ROPE to be the interval $[-0.01, 0.01]$. The results from Table \ref{tab:ahe_results} are corroborated by the Bayesian analysis. \texttt{LL} shows a significantly better ER and a comparable RP with respect to \texttt{CL}. Expectedly, there is a trade-off between ER and RP: greater ER leads to lower RP, and vice-versa. Notwithstanding, relative to \texttt{CL}, \texttt{LL} is able to significantly improve ER while keeping a comparable (i.e., within the region of practical equivalence) RP. Maximising ER in this particular domain of application is important because the events of interest are disruptive. While \texttt{LL} shows a significantly worse RP relative to \texttt{AH} and \texttt{RG}, it compensates with a better ER.

\begin{figure}[!hbt]
    \centering
    \includegraphics[width=\textwidth, trim=0cm 0cm 0cm 0cm, clip=TRUE]{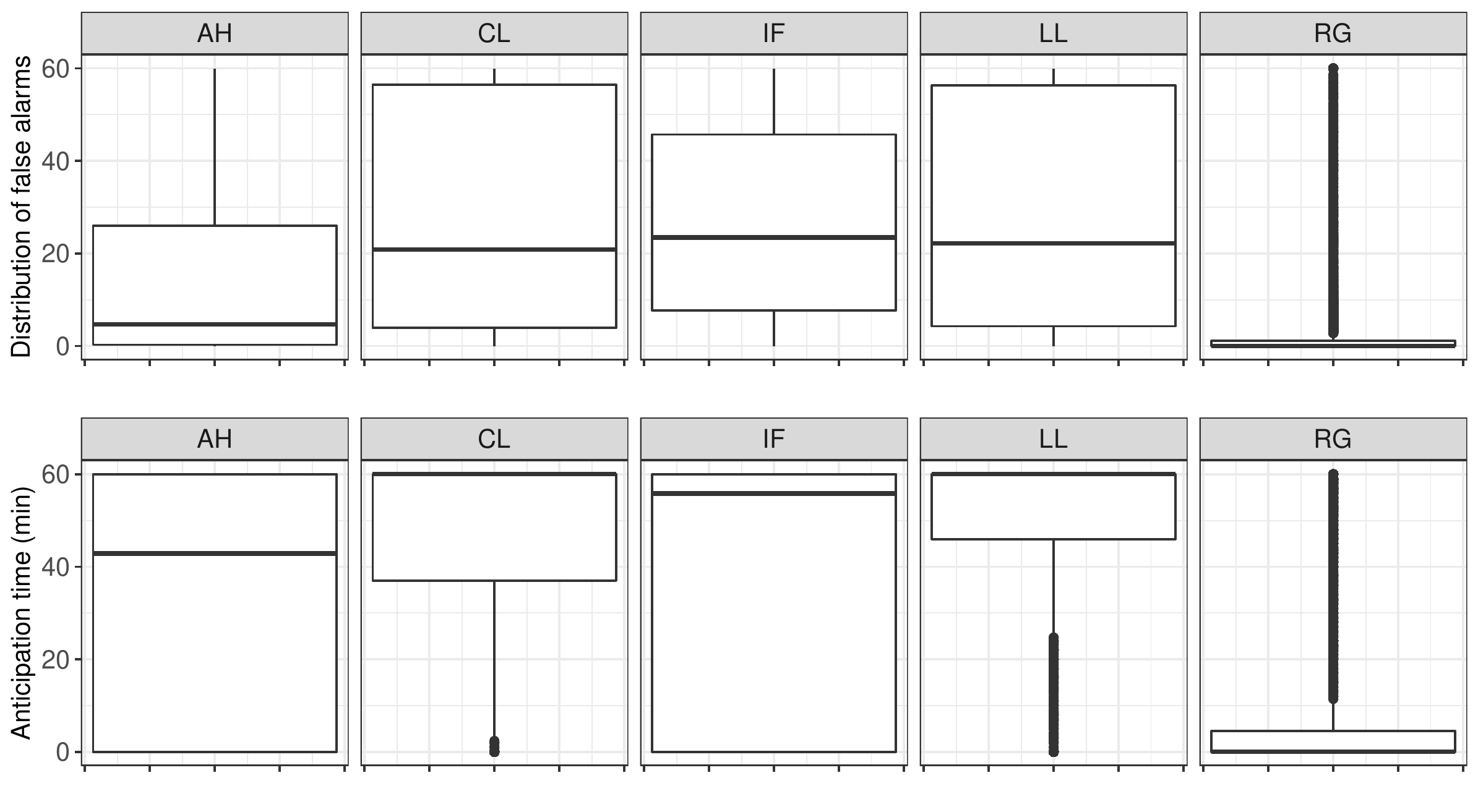}
    \caption{Distribution of the false alarms issued per hour and per patient (top), and the distribution of anticipation time (in minutes) per patient (low) for the AHE prediction problem}
    \label{fig:faph_ahe}
\end{figure}

In Figure \ref{fig:faph_ahe}, we show the distribution of the false alarms issued per hour and per patient (upper tile), and the distribution of anticipation time (in minutes) per patient (lower tiles), for each method under comparison. For instance, a value of 10 means that, on average, 10 false alarms are issued for a given patient (upper tile). On the other hand, a value of 30 in the distribution on the right side of the figure means that an AHE was predicted with 30 minutes in advance (before the episode started). The values of these distributions are somehow related to the previous results. \texttt{AH} and \texttt{RG} show the lowest average false alarms per hour, which correlates with the results of RP. Conversely, \texttt{LL} seems to have the most interesting distribution relative to anticipation time (close to the value of 60).

\subsection{Results on TE Prediction}

Similar to Table \ref{tab:ahe_results}, Table \ref{tab:te_results} presents the average of results, and respective standard deviation, across the 50 folds for the TE prediction problem. Overall, similar conclusions can be drawn from the performance metrics. The proposed method \texttt{LL} captures 93.8\% of the TE, which is a greater value relative to the remaining methods. Although slightly better, the RP value is comparable to \texttt{CL} (but worse than that of \texttt{AH} and \texttt{RG}). 

\begin{table}[!ht]
    \caption{Average of results for the TE prediction problem across the 50 folds}
    \label{tab:te_results}
    \begin{tabular}{l|r|r|r|r}
      \textbf{Method} & \textbf{ER} & \textbf{RP} & \textbf{Avg. AT} & \textbf{Avg. FA}\\
      \hline
      
      \texttt{AH} & 0.749$\pm$0.051 & \textbf{0.204}$\pm$0.022 & 37.9$\pm$3.1 & 6.9$\pm$2.7\\
      
      \texttt{CL} & 0.919$\pm$0.024 & 0.130$\pm$0.021 & 51.6$\pm$2.6 & 31.9$\pm$9.7\\
      
      \texttt{LL} & \textbf{0.938}$\pm$0.027 & 0.136$\pm$0.018 & \textbf{53.0}$\pm$2.2 & 35.9$\pm$8.8\\
      
      \texttt{IF} & 0.756$\pm$0.317 & 0.051$\pm$0.013 & 42.5$\pm$19.3 & 35.9$\pm$21.7\\
      
      \texttt{RG} & 0.646$\pm$0.049 & 0.195$\pm$0.025 & 31.1$\pm$2.8 &  \textbf{2.3}$\pm$0.9\\
    \end{tabular}%
\end{table}

\begin{figure}[!hbt]
    \centering
    \includegraphics[width=\textwidth, trim=0cm 0cm 0cm 0cm, clip=TRUE]{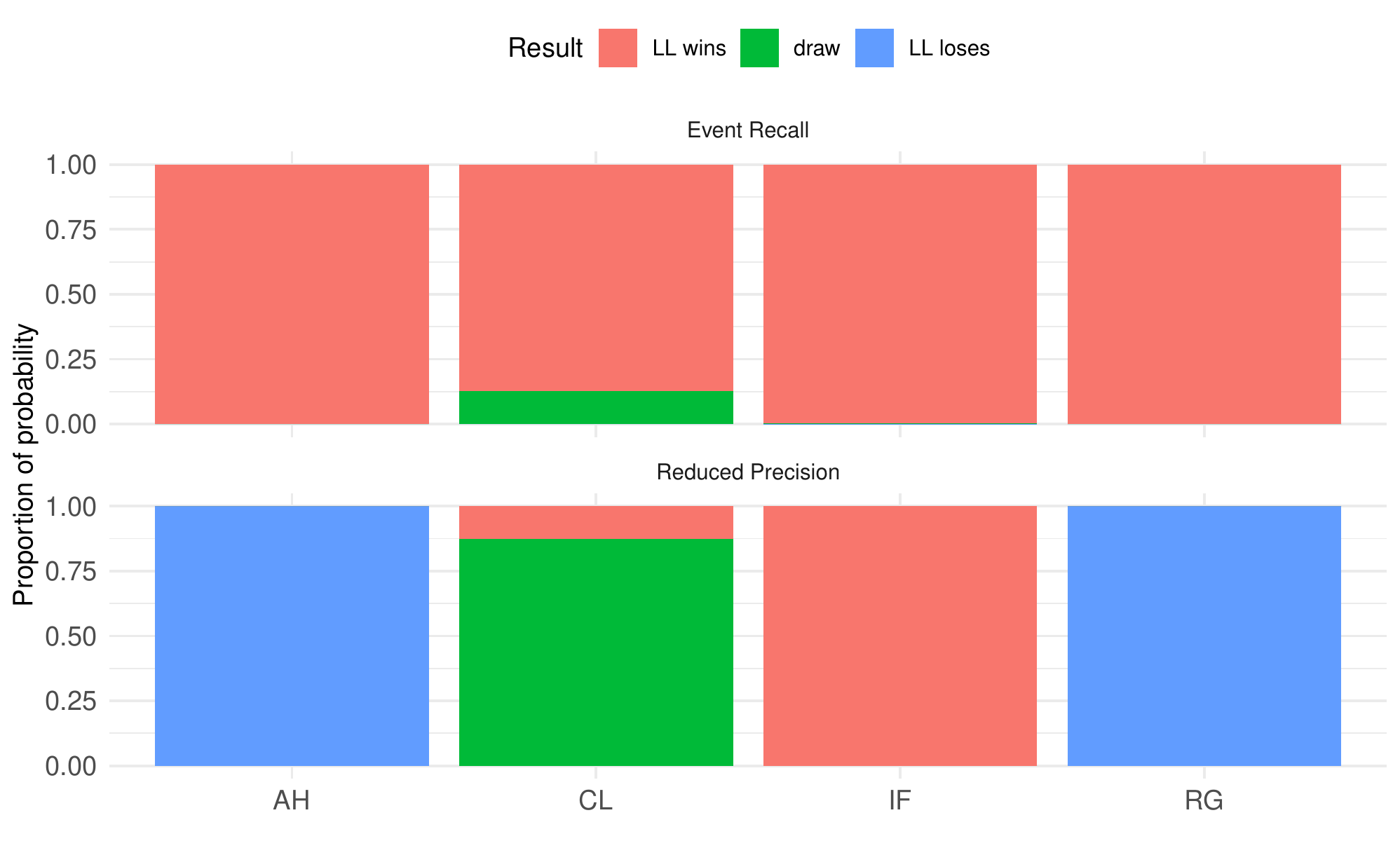}
    \caption{Comparing \texttt{CL} with \texttt{LL} with a Bayesian correlated t-test for ER and RP metrics (TE prediction)}
    \label{fig:bayes_te}
\end{figure}

This conclusion can also be drawn from the results of the Bayesian analysis shown in Figure \ref{fig:bayes_te}. \texttt{LL} shows a significantly better ER relative to the \texttt{CL} while having a comparable RP. \texttt{LL} also shows a better ER relative to the remaining approaches but loses to \texttt{AH} and \texttt{RG} when analysis RP.

\begin{figure}[!hbt]
    \centering
    \includegraphics[width=\textwidth, trim=0cm 0cm 0cm 0cm, clip=TRUE]{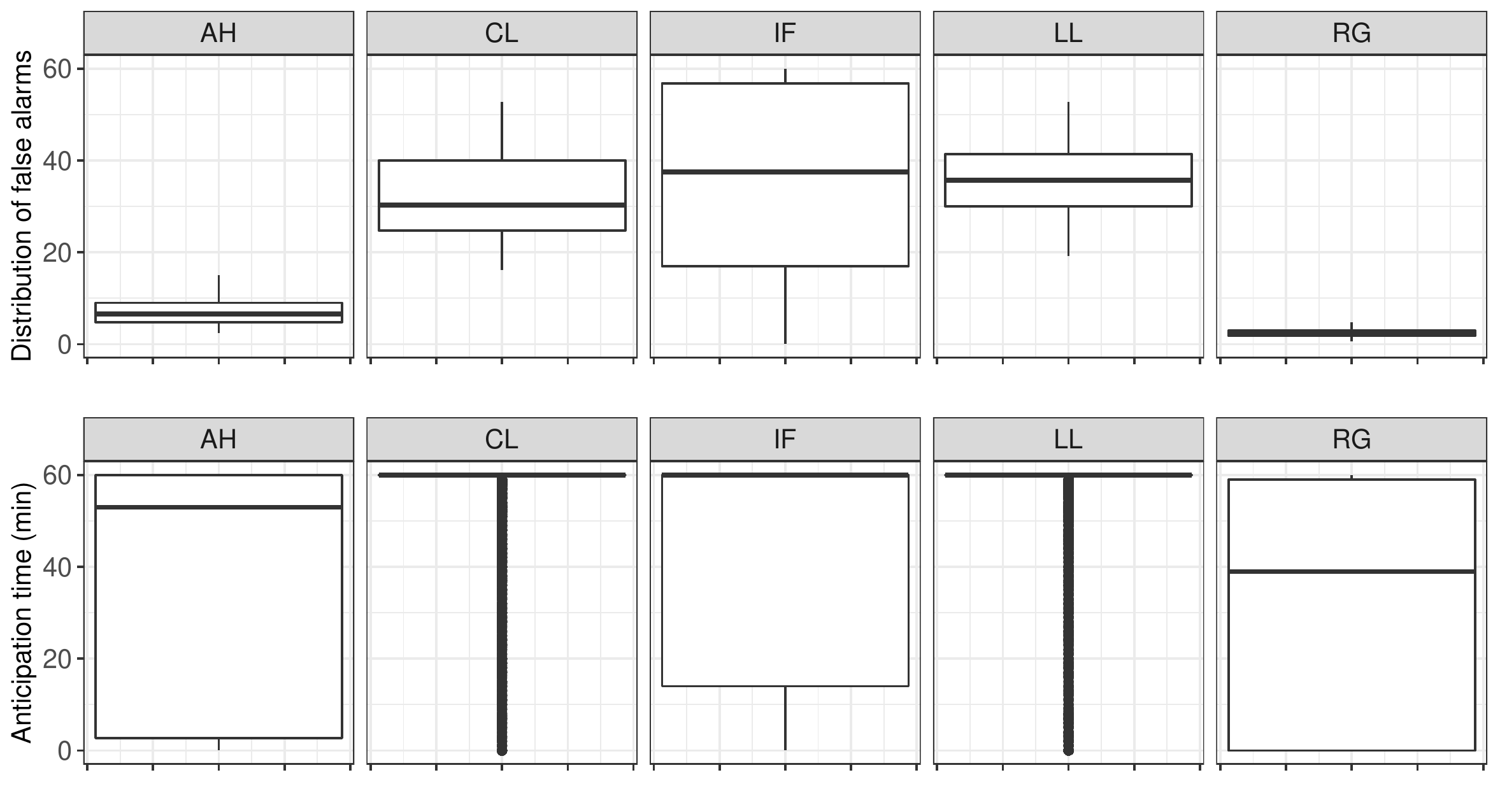}
    \caption{Distribution of the false alarms issued per hour and per patient (top), and the distribution of anticipation time (in minutes) per patient (low) for the TE prediction problem}
    \label{fig:faph_te}
\end{figure}

In Figure \ref{fig:faph_te}, we show the distribution of the false alarms issued per hour and per patient (left), and the distribution of anticipation time (in minutes) per patient (right), for each method under comparison. This analysis also shows a similar result and the same study for the AHE prediction problem shown in Figure \ref{fig:faph_ahe}.

\subsection{Performance by Layer}

We showed the competitiveness of \texttt{LL} relative to state of the art approaches to early anomaly detection in two case studies: AHE and TE prediction (\textbf{RQ1}). We address the research question \textbf{RQ2} in this section and analyse the predictive performance of each layer in the proposed layered learning architecture. We split this evaluation into the following three parts:

\begin{description}
    \item[\textbf{RQ2.1}] What is the performance of the model $f^\mathcal{S}$ (pre-conditional events sub-task)? That is, how well does the first layer of \texttt{LL} distinguish normal activity from the pre-conditional events  $\mathcal{S}^{\text{AHE}}$ or $\mathcal{S}^{\text{TE}}$ (c.f. right side of Figure~\ref{fig:LLDATASPACE})?
    \item[\textbf{RQ2.2}] Assuming there is an impending  pre-conditional event ($\mathcal{S}^{\text{AHE}}$ or $\mathcal{S}^{\text{TE}}$), what is the performance of the $f^\mathcal{F}$ model (main events sub-task)? In other words, how well does the second and final layer of \texttt{LL} is able to distinguish pre-conditional events from main events (AHE/TE)?
    \item[\textbf{RQ2.3}] Assuming that the first layer wrongly predicts that there is an impending pre-conditional event, i.e., a false positive in the pre-conditional events sub-task issued by $f^\mathcal{S}$. What is the performance of $f^\mathcal{F}$ (main events sub-task)? To be more precise, how well does the final layer of \texttt{LL} distinguish normal activity (but predicted to be a pre-conditional event by the first layer) from main events. 
\end{description}

To answer these questions, we use standard binary classification metrics, namely recall, precision, F-score, and specificity. Note that in the previous subsections, we were concerned with how well each method captured the events of interest (AHE and TE). To evaluate this, we used appropriate measures (ER and RP). In this analysis, however, we want to understand the ability of each layer in \texttt{LL} to distinguish the different scenarios in each sub-sequence. The results are presented in Table~\ref{tab:ahe_layers} for AHE prediction, and Table~\ref{tab:te_layers} for TE prediction.

\begin{table}[bht]
    \caption{Performance of each component in the proposed layered learning architecture for the AHE prediction problem.}
    \label{tab:ahe_layers}
    \begin{tabular}{l|r|r|r|r}
      \textbf{Analysis} & \textbf{Recall} & \textbf{Precision} & \textbf{F-score} & \textbf{Specificity}\\
      \hline
      \textbf{RQ2.1} & 0.769$\pm$0.038 & 0.391$\pm$0.035 & 0.517$\pm$0.030 & 0.761$\pm$0.043\\
      \textbf{RQ2.2} & 0.683$\pm$0.103 & 0.290$\pm$0.062 & 0.402$\pm$0.064 & 0.595$\pm$0.108\\
      \textbf{RQ2.3} & NA & NA & NA & 0.524$\pm$0.126\\
    \end{tabular}%
\end{table}

\begin{table}[bht]
    \caption{Performance of each component in the proposed layered learning architecture for the TE prediction problem.}
    \label{tab:te_layers}
    \begin{tabular}{l|r|r|r|r}
      \textbf{Analysis} & \textbf{Recall} & \textbf{Precision} & \textbf{F-score} & \textbf{Specificity}\\
      \hline
      \textbf{RQ2.1} & 0.890$\pm$0.039 & 0.476$\pm$0.069 & 0.618$\pm$0.061 & 0.844$\pm$0.034\\
      \textbf{RQ2.2} & 0.821$\pm$0.278 & 0.616$\pm$0.065 & 0.670$\pm$0.163 & 0.188$\pm$0.289\\
      \textbf{RQ2.3} & NA & NA & NA & 0.183$\pm$0.281\\
    \end{tabular}%
\end{table}

The model $f^\mathcal{S}$ presents a reasonable performance for capturing pre-conditional events, with an average F-score of 0.517 and 0.618, for AHE prediction and TE prediction, respectively (\textbf{RQ2.1}). The model $f^\mathcal{F}$ presents an average F-score of 0.402 and 0.670 for capturing main events inside the pre-conditional events data space (\textbf{RQ2.2}). While the recall of both these components seems adequate, the precision is lower than expected. 

One of the main challenges behind using the proposed layered learning architecture is that errors may propagate from layer to layer. We evaluate one scenario where this might occur: when the model for pre-conditional events $f^\mathcal{S}$ issues a false alarm. That is, when it predicts that there is an impending pre-conditional event when in fact there is not (data remains as normal activity). We analyse how the second model for capturing main events $f^\mathcal{F}$ performs in such conditions. Ideally, $f^\mathcal{F}$ should ignore (i.e. classify as negative) all these observations.

The value of specificity averages at 0.524 and 0.183, for AHE prediction and TE prediction, respectively. These values represent the ratio of sub-sequences, which the model $f^\mathcal{F}$ can correctly identify a non-main event, after the first model $f^\mathcal{S}$ has issued a false alarm. Note that the true value is always negative, so only the specificity metric makes sense in this case. While the overall layered architecture generally performs better than the typical approach (\texttt{CL}), we believe that these results show that there is room for improvement.

\subsection{Run-time Analysis}

In the previous sections, we analysed \texttt{LL} in terms of predictive performance. In this section, we address the research question \textbf{RQ3} by analysing \texttt{LL} in terms of computation time. To accomplish this, we measure the time spent in fitting each method and using it to predict the test set of a given cross-validation fold.

\begin{table}[!ht]
    \caption{Average run-time in seconds, and respective standard deviation, of each method across the 50 folds}
    \centering
    \label{tab:rt}
    \begin{tabular}{l|r|r}
      \textbf{Method} & \textbf{AHE prediction} & \textbf{TE prediction} \\
      \hline
      
      \texttt{AH} & \textbf{0} & \textbf{0}\\
      
      \texttt{CL} & 15.6$\pm$2.5 & 7.4$\pm$0.8\\
      
      \texttt{LL} & 102.3$\pm$12.5 & 70.9$\pm$4.9\\
      
      \texttt{IF} & 67.3$\pm$7.5 & 53.3$\pm$2.6\\
      
      \texttt{RG} & 416.3$\pm$30.1 & 344.2$\pm$21.7\\
    \end{tabular}%
\end{table}

Table \ref{tab:rt} shows the average run-time in seconds, and respective standard deviation, of each method across the 50 folds. The relative results are similar for both AHE prediction, and TE prediction. \texttt{LL} takes, on average, more than one minute to run. Although this value is not considerable, both \texttt{CL} and \texttt{IF} take less time to compute than \texttt{LL}. The regression-based method \texttt{RG} is the one that takes more time to compute. This is expected since the underlying model is a multi-step forecasting method -- a learning model is created for each point in the target window (i.e. 30 models), which significantly drives the run-time of this approach. Finally, \texttt{AH} is not a machine learning method. It is a rule derived from domain expertise which issues alarms when a determined variable exceeds some value. In effect, we consider the run-time of \texttt{AH} to be negligible.

\subsection{Re-sampling Analysis}\label{sec:resampleanalysis}

In this section, we address the research question \textbf{RQ4}. As we mentioned before, the proposed method \texttt{LL} and the state of the art approach \texttt{CL} are applied after pre-processing the data set with a re-sampling method. Re-sampling strategies are commonly used to mitigate the class imbalance problem \cite{branco2016survey}, including in CHE prediction problems \cite{lee2010hypotensive,forkan2017visibid}. We analyse the impact of several re-sampling strategies in terms of ER and RP. In this analysis, we focus on the methods \texttt{LL} and \texttt{CL}. The methods \texttt{IF} and \texttt{AH} do not require balancing the distribution. To our knowledge, \texttt{RG} has been applied to CHE prediction without such procedures \cite{lee2010hypotensive,rocha2011prediction}.

We tested the following six different strategies:
\begin{itemize}
    \item No re-sampling (NR), in which the distribution is left imbalanced;
    \item Random Under-sample (RU): in this strategy cases from the majority class are randomly removed until the distribution of classes is balanced;
    \item Random Over-sample (RO): Similarly to RU, in a RO approach random instances from the minority class are replicated the distribution of classes is balanced;
    
    \item SMOTE (Synthetic Minority Over-sampling Technique) \cite{chawla2002smote}: instead of replicating instances from the minority class, SMOTE generates new synthetic observations similar to these. This is achieved by interpolation from a number of nearest neighbours;
    
    \item ADASYN (Adaptive Synthetic) \cite{he2008adasyn}: this method is another over-sampling technique which is similar to SMOTE. The core distinction is that ADASYN focuses on instances from the minority class which are more \textit{difficult to learn}, i.e., closer to the decision boundary;
    
    \item TOMEK: Tomek links is an under-sampling method \cite{tomek1976two}. It works by finding pairs of observations which are the nearest neighbour of each other, but of different classes. One can then remove the instance from the majority class, or even the respective pair \cite{batista2004study}. We adopt the latter approach.
\end{itemize}

Table \ref{tab:ahe_ra_results} shows the result of the analysis for the AHE prediction problem. Almost all the re-sampling approaches improve the sensitivity of the methods for the event of interest (i.e. higher ER). The exception is TOMEK for \texttt{LL}, and ADASYN for \texttt{CL}. However, there is a trade-off with RP, which generally decreases with the application of the re-sampling methods. Since in this particular domain of application we are focused on preventing CHEs, we choose the re-sampling method emphasising ER. This justifies the pick of a RU approach for \texttt{LL}, and the choice for RO for \texttt{CL}.

\begin{table}[!ht]
  \centering
    \caption{Results of the re-sampling analysis for the AHE problem (average across the 50 folds)}
    \label{tab:ahe_ra_results}
    \begin{tabular}{r|l|l|l|l}
    
        & \multicolumn{2}{c|}{\texttt{LL}} & \multicolumn{2}{c|}{\texttt{CL}} \\
        \hline
    
      \textbf{Method} & \textbf{ER} & \textbf{RP} & \textbf{ER} & \textbf{RP}\\
      \hline
      
      NR & 0.778$\pm$0.07 & \textbf{0.107}$\pm$0.02 & 0.755$\pm$0.08 & 0.108$\pm$0.02\\
      
      RU & \textbf{0.830}$\pm$0.05 & 0.090$\pm$0.02 & 0.792$\pm$0.07 & 0.091$\pm$0.02\\
      
      RO & 0.828$\pm$0.06 & 0.087$\pm$0.02 & \textbf{0.807}$\pm$0.07 & 0.089$\pm$0.02\\
      
      SMOTE & 0.829$\pm$0.06 & 0.083$\pm$0.02 & 0.805$\pm$0.06 & 0.085$\pm$0.02\\
      
      ADASYN & 0.788$\pm$0.07 & 0.105$\pm$0.03 & 0.730$\pm$0.08 & \textbf{0.112}$\pm$0.03\\
      
      TOMEK & 0.774$\pm$0.07 & 0.101$\pm$0.03 & 0.767$\pm$0.09 & 0.098$\pm$0.03\\
      
    \end{tabular}%
\end{table}

A similar analysis can be made for TE prediction, whose results are shown in Table \ref{tab:te_ra_results}. In this case, both methods (\texttt{LL} and \texttt{CL}) present their best results when applied with the SMOTE re-sampling strategy.

\begin{table}[!ht]
  \centering
    \caption{Results of the re-sampling analysis for the TE problem (average across the 50 folds)}
    \label{tab:te_ra_results}
    \begin{tabular}{r|l|l|l|l}
    
        & \multicolumn{2}{c|}{\texttt{LL}} & \multicolumn{2}{c|}{\texttt{CL}} \\
        \hline
    
      \textbf{Method} & \textbf{ER} & \textbf{RP} & \textbf{ER} & \textbf{RP}\\
      \hline
      
      NR & 0.886$\pm$0.04 & \textbf{0.165}$\pm$0.02 & 0.853$\pm$0.04 & 0.169$\pm$0.03\\
      
      RU & 0.924$\pm$0.03 & 0.141$\pm$0.02 & 0.903$\pm$0.04 & 0.138$\pm$0.02\\
      
      RO & 0.930$\pm$0.03 & 0.142$\pm$0.02 & 0.900$\pm$0.05 & 0.147$\pm$0.02\\
      
      SMOTE & \textbf{0.938}$\pm$0.03 & 0.136$\pm$0.02 & \textbf{0.919}$\pm$0.06 & 0.130$\pm$0.02\\
      
      ADASYN & 0.887$\pm$0.03 & \textbf{0.165}$\pm$0.02 & 0.815$\pm$0.06 & \textbf{0.179}$\pm$0.03\\
      
      TOMEK & 0.893$\pm$0.05 & 0.149$\pm$0.03 & 0.838$\pm$0.05 & 0.165$\pm$0.03\\
      
    \end{tabular}%
\end{table}

\section{Discussion}\label{chp6:sec:discussion}

\subsection{On the Experimental Results}

In the previous section, we provided empirical evidence for the advantages of using a layered learning approach for CHE prediction problems. We briefly discuss the results in this section. We also discuss the main challenges associated with the proposed approach.

\texttt{IF}, a state of the art approach to anomaly detection, shows a significantly worse performance relative to discriminating approaches, namely \texttt{LL} and \texttt{CL}, for both AHE and TE prediction problems. A regression approach (\texttt{RG}) also shows a significantly lower ER with respect to the other methods. It shows the highest RP, which indicates that this type of approach is conservative (low recall of events and low average number of false alarms). In summary, \texttt{LL} shows a competitive performance relative to state of the art approaches to solve early anomaly detection predictive tasks.

As we mentioned before, the reported experiments were carried out using an extreme gradient boosting learning algorithm \cite{chen2015xgboost}. This algorithm was used to train both layers of our approach (\texttt{LL}), and as a stand-alone classifier without layered learning (\texttt{CL}). Using this learning algorithm lead to the best overall results relative to other ones such as random forests, or a deep feedforward neural network. Notwithstanding, deep learning approaches, recurrent architectures in particular, have been increasingly applied in the healthcare domain (e.g. \cite{tamilselvan2013failure}). In future work, we will study these methods further, both as benchmarks and as possible solutions within a layered learning approach.

\subsection{Future Work}

\subsubsection{Towards the Automatic Definition of Pre-conditional Events}\label{sec:towards}

One of the main challenges in the proposed methodology is the manual definition of pre-conditional events. This process is highly domain-dependent. In this sense, it can be regarded as an opportunity for domain experts to embed their expertise in predictive models. Notwithstanding, nowadays there is an increasing interest for end-to-end automated machine learning technologies \cite{thornton2013auto,feurer2015efficient}, and a manual definition of sub-tasks can be regarded as a bottleneck. 

The problem of manually defining sub-tasks is common in other hierarchical approaches. For example, similarly to layered learning, hierarchical reinforcement learning also involves the decomposition of a problem into hierarchical sub-tasks \cite{dietterich2000hierarchical}. In this topic, one of the most common approaches to this effect is the options framework by Sutton et al \cite{sutton1998between}. According to this approach, the definition of the sub-tasks in performed manually by the programmer.

An automatic definition of sub-tasks, which in our case refers to the definition of pre-conditional events, is a difficult problem. In the reinforcement learning literature, there are several recent works which try to learn these sub-tasks \cite{klissarov2017learnings,harb2018waiting,riemer2018learning}. In future work, we will explore this research line and try to leverage it to develop a way of automatically defining pre-conditional events.

\subsubsection{Other Research Lines}

Although we focus on CHE prediction problems, our ideas for layered learning can be generally applied to other early anomaly detection or activity monitoring problems, for example, problems with complex targets, which can be decomposed into partial, simpler targets. While the task decomposition is dependent on the domain, we describe some guidelines which can facilitate its implementation. 

In Section \ref{sec:ss4tr}, we described the pipeline for formalizing the predictive task, i.e., how to transform the set of time series into a geometrical form for learning a predictive model. As we mentioned, we settled on an approach recommended by previous work. Notwithstanding, studying the most appropriate way of formalizing the predictive task (from the set of time series to a geometrical form).

\section{Related Research}\label{chp6:sec:relatedwork}

\subsection{Activity Monitoring}\label{sec:am}

The problem of timely detection of anomalies is also known in the literature as activity monitoring \cite{fawcett1999activity}. The goal of this predictive task is to track a given activity over time and launch timely alarms about interesting events that require action.  
According to Fawcett and Provost \cite{fawcett1999activity}, there are two classes of methods for activity monitoring:
\begin{itemize}
	\item Profiling: In a profiling strategy, a model is constructed using only the normal activity of the data, without reference to abnormal cases. Consequently, an alarm is triggered if the current activity deviates significantly from normal activity. This approach may be useful in complex time-dependent data where anomalies do not have a well-defined concept. For example, fraud attempts often occur in different manners. Effectively, by modelling only normal activity, one is apt to detect different types of anomalies, including the ones unknown hitherto.
	\item Discriminating: A discriminating method constructs a model about anomalies with respect to the normal activity, handling the problem as a classification one. A system then uses a model to examine the time series and look for anomalies. In this scenario, the recent past dynamics of the data are used as predictor variables. The target variable denotes whether the event of interest occurs.
\end{itemize}

We focus on the latter strategy, which is the one followed by the proposed layered learning method for activity monitoring. Notwithstanding, we compare our approach to \texttt{IF}, which is a method that follows the profiling strategy.

\subsection{CHE Prediction}

AHE prediction has been gaining increasing attention from the scientific community. For example, the $10^{th}$ annual PhysioNet / Computers in Cardiology Challenge focused on this predictive task \cite{moody2009predicting}. While the methods used in this particular challenge are not state of the art anymore, the purpose of the reference is to show the relevance of the predictive task.

Like other activity monitoring problems or anomaly detection tasks, the typical approach to this problem is to use standard classification methods. This is the case of Lee and Mark, which use a feed-forward neural network as predictive model \cite{lee2010investigation}. Tsur et al. \cite{tsurhypotensive} follow a similar approach and also propose an enhanced feature extraction approach before applying an extreme gradient boosting algorithm. In turn, Rocha et al. \cite{rocha2011prediction} propose a regression approach by forecasting future values of blood pressure. In their approach, alarms for impending AHE are launched according to a deterministic function which receives as input the numeric predictions. TE prediction also is a relevant task. For example, Forkan et al. \cite{forkan2017visibid} propose a predictive model for detecting several health conditions, including tachycardia and hypotension. 

\subsection{Layered Learning}

Layered learning was proposed by Stone and Veloso \cite{stone2000layered}, and was specifically designed for scenarios with a complex mapping from inputs to outputs. In particular, they applied this approach to improve several processes in robotic soccer. 

Decroos et al. \cite{decroos2017predicting} apply a similar approach for predicting goal events in soccer matches. Instead of directly modelling such events, they first model goal attempts as what we call in this paper as a pre-conditional events sub-task.

Layered learning stems from the more general topic of multi-strategy learning. Layered learning approaches run multiple learning processes to improve the generalisation in a predictive task. This is a similar strategy as ensemble learning methods, which we used in the previous part of this thesis. The main difference is that in layered learning, each layer addresses a different predictive task, while in ensemble learning the predictive task is typically a single one.

\subsection{Related Early Decision Systems}

The need for early predictions is also important in other predictive tasks which are related to activity monitoring. Time series classification is a well-studied topic, for example, in data streams mining \cite{bifet2009data}. However, traditional time series classification methods are inflexible for early classification. Typically, a method is trained on the full length of the time series, and the prediction is also made at that time point. Therefore, the main limitation of such methods is that they ignore the sequential nature of data, and the importance of \textit{early} classification \cite{fawcett1999activity}. The earliness component of classifiers for time series is important so that professionals and decision-makers can take pro-active measures and timely decisions. To overcome this limitation, several models for early classification of time series have been proposed. Some examples are the works of He et al. \cite{he2015early}, or Xing et al. \cite{xing2011extracting}. Another example of a type of early decision systems is human motion recognition \cite{kuehne2011hmdb}. This task is fundamental for surveillance systems or human-computer interactive systems.

\section{Summary}\label{chp6:sec:conclusions}

Layered learning approaches are designed to solve predictive tasks in which a direct mapping from inputs to outputs is difficult. In this paper, we developed a layered learning approach for the early detection of anomalies in time series data. The idea is to break the original predictive task into two simpler predictive tasks, which are, in principle, easier to solve. We create an initial model that is designed to distinguish normal activity from a relaxed version of anomalous behaviour (pre-conditional events). A subsequent model is created to distinguish such pre-conditional events from the actual events of interest.

We have focused on predicting critical health conditions in ICUs, namely hypotension and tachycardia events. Compared to standard classification, which is a common solution to this type of predictive tasks, the proposed model can capture significantly more anomalous events with a comparable number of false alarms. The results also suggest that the proposed approach is better than other state of the art methods.


\begin{acknowledgements}

This work was financed by the Portuguese funding agency, FCT - Fundação para a Ciência e a Tecnologia, through national funds, and co-funded by the FEDER, where applicable.

\end{acknowledgements}

\bibliographystyle{spmpsci}      

\end{document}